\documentclass[lettersize,journal]{IEEEtran}
\usepackage{graphics} % for pdf, bitmapped graphics files
\usepackage{epsfig} % for postscript graphics files
\usepackage{mathptmx} % assumes new font selection scheme installed
\usepackage{times} % assumes new font selection scheme installed
\usepackage{amsmath} % assumes amsmath package installed
\usepackage{amssymb}  % assumes amsmath package installed
\usepackage{multirow}
\usepackage{amsfonts}
\usepackage{booktabs}
\usepackage[table,xcdraw]{xcolor}
\usepackage{cite}
\usepackage{url}
\usepackage{threeparttable}
\usepackage{bigdelim}
\usepackage{indentfirst}
\usepackage{colortbl}
\usepackage{bbding}
\usepackage{subfiles}
\usepackage[pagebackref=true,breaklinks=true, colorlinks, bookmarks=false]{hyperref}

\usepackage{bm}
\usepackage{makecell}
\usepackage{orcidlink}
\usepackage{pifont}% http://ctan.org/pkg/pifont
\usepackage{newtxtext, newtxmath}
\newcommand{\cmark}{\ding{51}}%
\newcommand{\xmark}{\ding{55}}%
\usepackage{xspace}
\newcommand{\ie}{{\emph{i.e.}},\xspace}
\newcommand{\eg}{{\emph{e.g.}},\xspace}
\newcommand{\etc}{etc.}

\newcommand{\vs}{{\emph{vs.}}\xspace}

\begin{document}

\title{
DQFormer: Towards Unified LiDAR Panoptic Segmentation with Decoupled Queries
}

\author{
Yu Yang~\orcidlink{0000-0001-9292-6932}, Jianbiao Mei~\orcidlink{0000-0003-3849-2736}, Liang Liu~\orcidlink{0000-0001-7910-810X}, Siliang Du, Yilin Xiao~\orcidlink{0000-0002-9827-6149}, \\
Jongwon Ra~\orcidlink{0009-0005-8815-5350}, Yong Liu~\orcidlink{0000-0003-4822-8939}, \emph{Member, IEEE}, Xiao Xu, and Huifeng Wu~\orcidlink{0000-0002-7189-1205}, \emph{Member, IEEE}% <-this % stops a space
\thanks{This work was supported by a Grant from The National Natural Science Foundation of China (No. U21A20484). \emph{(Equal Contribution: Yu Yang and Jianbiao Mei.)} \emph{(Corresponding Author: Yong Liu and Xiao Xu.)}}
\thanks{
Yu Yang, Jianbiao Mei, Liang Liu, Jongwon Ra, and Yong Liu are with the Institute of Cyber-Systems and Control, Zhejiang University, Hangzhou, China (e-mail: yu.yang@zju.edu.cn; jianbiaomei@zju.edu.cn; leonliuz@zju.edu.cn; jongwonra@zju.edu.cn; yongliu@iipc.zju.edu.cn).}
\thanks{Siliang Du, and Yilin Xiao are with Huawei Technologies Co., Ltd., Wuhan, China (e-mail: dusi@whu.edu.cn; xiaoyilin@whu.edu.cn).}
\thanks{Xiao Xu is with the Institute of Industrial Technology Research, Zhejiang University, Hangzhou, China (e-mail: xuxiao0224@126.com).}
\thanks{Huifeng Wu is with the Institute of Intelligent and Software Technology, Hangzhou Dianzi University, Hangzhou, China (e-mail: whf@hdu.edu.cn).}
}
\markboth{Journal of \LaTeX\ Class Files,~Vol.~14, No.~8, August~2022}%
{Shell \MakeLowercase{\textit{et al.}}: A Sample Article Using IEEEtran.cls for IEEE Journals}


\maketitle

\begin{abstract}
LiDAR panoptic segmentation, which jointly performs instance and semantic segmentation for \emph{things} and \emph{stuff} classes, plays a fundamental role in LiDAR perception tasks. 
While most existing methods explicitly separate these two segmentation tasks and utilize different branches (i.e., semantic and instance branches), some recent methods have embraced the query-based paradigm to unify LiDAR panoptic segmentation. 
However, the distinct spatial distribution and inherent characteristics of objects(things) and their surroundings(stuff) in 3D scenes lead to challenges, including the mutual competition of things/stuff and the ambiguity of classification/segmentation.
In this paper, we propose decoupling things/stuff queries according to their intrinsic properties for individual decoding and disentangling classification/segmentation to mitigate ambiguity. 
To this end, we propose a novel framework dubbed DQFormer to implement semantic and instance segmentation in a unified workflow. 
Specifically, we design a decoupled query generator to propose informative queries with semantics by localizing things/stuff positions and fusing multi-level BEV embeddings. 
Moreover, a query-oriented mask decoder is introduced to decode corresponding segmentation masks by performing masked cross-attention between queries and mask embeddings. 
Finally, the decoded masks are combined with the semantics of the queries to produce panoptic results.
Extensive experiments on nuScenes and SemanticKITTI datasets demonstrate the superiority of our DQFormer framework.
\end{abstract}


\section{Introduction}
\IEEEPARstart{W}{ith} the rapid development of autonomous driving and robotics in recent years, LiDAR perception has gained significant attention in the community. LiDAR segmentation tasks involve point-level predictions of the entire scene, which are fundamental and critical for perception tasks. LiDAR Panoptic Segmentation (LPS) not only predicts the point-wise semantic label for \emph{stuff} classes (e.g., road and vegetation) but also the label and instance IDs for \emph{thing} classes (e.g., car and person). Due to its ability to unify semantic and instance segmentation in a single architecture, LPS plays an essential role in LiDAR understanding and has a wide range of applications.

\begin{figure}[t]
\centering
	\includegraphics[width=0.49\textwidth]{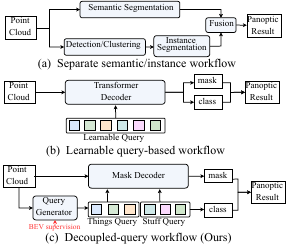}
    \vspace{-2em}
	\caption{(a) Existing semantic/instance separation paradigm. (b) Existing learnable query-based methods ignore the distinctions between things and stuff. (c) We decouple things/stuff queries and mitigate competition between classification/segmentation for unified LiDAR panoptic segmentation.}
	\label{fig:teaser2}
    \vspace{-1.5em}
\end{figure}

Most existing LPS methods \cite{hong2021lidar, sirohi2021efficientlps, zhou2021panoptic, li2022PHNet, mei2023centerlps} explicitly separate semantic/instance segmentation tasks and utilize two branches to implement panoptic segmentation. As illustrated in Figure \ref{fig:teaser2}(a), the semantic branch predicts the semantic label for each point, while the instance branch adopts segmentation-by-detection or regression-then-clustering paradigms to group instance IDs.
Inspired by the recent success of query-based methods in the 2D segmentation domain \cite{li2021fully, cheng2021per, cheng2022masked, zhang2021k, li2022panoptic}, MaskPLS \cite{marcuzzi2023mask} and PUPS \cite{su2023pups} proposed to use a set of learnable queries to implement unified LPS. This approach predicts a set of non-overlapping binary masks and semantic classes for either a stuff class or a potential object, as shown in Figure \ref{fig:teaser2}(b).

\begin{figure}[t]
\centering
	\includegraphics[width=0.49\textwidth]{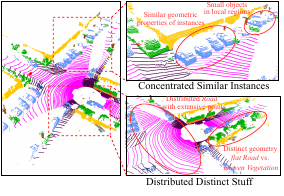}
    \vspace{-2.5em}
	\caption{Distinction between \emph{things} and \emph{stuff} in LiDAR scenes: Instances with similar geometric properties are typically concentrated in local regions, whereas distributed stuff with extensive points exhibit distinct geometries.}
	\label{fig:teaser1}
    \vspace{-1.5em}
\end{figure}

However, directly transferring commonly used query-based methods to LPS such as MaskPLS \cite{marcuzzi2023mask} and PUPS \cite{su2023pups} ignores the notable distinction between \emph{things} and \emph{stuff} in 3D scenarios as shown in Figure \ref{fig:teaser1}.
1) \emph{Disparate spatial distribution:} Stuff is typically distributed throughout the entire scene (\,  e.g., roads and vegetation) and constitutes a larger proportion of the point cloud. In contrast, instances are generally concentrated within specific local regions and are significantly smaller relative to the whole scene. 
2) \emph{Divergent inherent characteristics:} Various stuff classes exhibit distinct geometric attributes (\eg flat road surfaces \vs uneven vegetation points), which can serve as valuable distinguishing features in semantic classification for stuff points.
Conversely, instances of the same category share similar geometric properties while lacking distinctive texture or color characteristics, complicating the instance segmentation.

Due to the challenges posed by point clouds' intrinsic properties, vanilla query-based methods face substantial problems when applied to LiDAR scenes.
1) \emph{Mutual competition between things and stuff:} Learnable queries equipped with the Hungarian matching algorithm empirically tend to respond to a relatively large area (\ie stuff classes) for high recall, posing a particular challenge when segmenting smaller instances.
2) \emph{Ambiguity between classification and segmentation:} The vanilla query-based methods simultaneously predict categories and binary masks based on the learnable queries, inevitably leading to ambiguity between instances. We explain that classification supervision promotes feature representations to be more similar between different objects of the same category, thus complicating the segmentation between different instances. 

Thus, based on the above observations, we propose decoupling things/stuff queries according to their intrinsic properties for individual decoding and disentangling classification/segmentation to mitigate ambiguity.  
As shown in Figure \ref{fig:teaser2}(c), we design a query generator to produce two types of queries with assigned semantic classes, \ie things and stuff queries. 
Unlike the commonly used learnable manner, the query generator produces informative queries based on the positions and embeddings of things/stuff in multi-level BEV (Bird's Eye View) space. Our main insights include: (1) Localizing \emph{things} in BEV is more convenient and efficient, facilitating the generation of distinct instance queries; (2) The receptive field can be large enough while using limited computation overhead in BEV for \emph{stuff}. Additionally, we propose a query-oriented mask decoder that predicts the segmentation masks based on the embeddings of things and stuff queries. These masks are combined with the semantic classes of queries to produce panoptic results. Our decoupled pipeline mitigates the conflicts between things/stuff and classification/segmentation while maintaining a unified workflow.

More specifically, with the proposed query generator and mask decoder, we construct a novel framework termed \textbf{DQFormer} to implement semantic and instance segmentation in one workflow. DQFormer comprises a multi-scale feature encoder, a decoupled query generator, and a query-oriented mask decoder. 
The feature encoder first extracts voxel-level features and point-wise embeddings of multi-resolutions. Then, to accurately locate objects of varying sizes and identify small instances, the query generator produces query proposals from multi-level BEV embeddings and leverages a query-fusion module to merge these proposals into informative queries with multi-scale receptive fields.
Once equipped with queries containing object features, the query-oriented mask decoder employs a masked cross-attention mechanism to decode the corresponding masks.

We evaluate our method on nuScenes \cite{caesar2020nuscenes} and SemanticKITTI \cite{behley2019semantickitti}, and the results demonstrate the effectiveness of our method. Our main contributions are as follows:

\begin{itemize}
\item We propose a novel framework dubbed DQFormer, which decouples things/stuff queries and disentangles classification/segmentation for unified LiDAR panoptic segmentation.

\item We design a multi-scale query generator that proposes semantic-aware queries by localizing things/stuff positions and fusing multi-level BEV embeddings. 

\item We propose a query-oriented mask decoder that utilizes informative queries to guide the segmentation process through a masked cross-attention mechanism.

\item Extensive experiments on nuScenes and SemanticKITTI show that our DQFormer achieves superior performance.
\end{itemize}

\begin{figure*}[t]
\centering
	\includegraphics[width=0.95\textwidth]{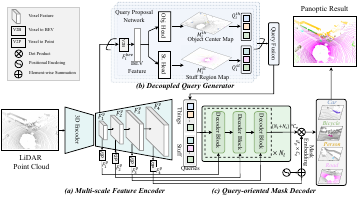}
 \vspace{-1em}
	\caption{\textbf{Overview of DQFormer.} (a) The feature encoder is applied to extract voxel features and point embeddings at multi-resolutions. (b) The query generator is designed to produce informative things/stuff queries with assigned semantics according to their positions and embeddings in BEV space. (c) The mask decoder performs masked cross-attention between queries and multi-level point embeddings to decode segmentation masks. Finally, the decoded masks are combined with the semantics of the queries to produce the panoptic result. Details of the decoder block are illustrated in Figure \ref{fig:decoder}.}
	\label{fig:pipeline}
 \vspace{-1em}
\end{figure*}

\section{Related Work}
\subsection{LiDAR Panoptic Segmentation}
Most existing LPS methods consist of separate semantic and instance branches. In terms of instance segmentation implementation, existing methods can be categorized into three types of frameworks: detection-based, clustering-based, and center-based methods. Some recent approaches propose to unify LPS in a learnable query-based manner. 

\noindent\textbf{Detection-based methods} \cite{hurtado2020mopt, sirohi2021efficientlps, xu2023aop, ye2022lidarmultinet, agarwalla2023lidar, xu2023aop} employ a 3D detection network \cite{yin2021center, zhou2022centerformer, fan2022embracing} to detect the bounding box of objects and then extract thing points located in the bbox to generate instance masks. 
Although the existing detection-based methods explicitly predict the position and size of the instances, the semantic branch is still indispensable for extracting thing points from boxes.

\noindent\textbf{Clustering-based methods} \cite{milioto2020lidar, li2022smac, hong2021lidar, zhou2021panoptic, li2022PHNet, gasperini2021panoster, li2023center, mei2023panet, nunes2022unsupervised, zhao2022divide, zhang2023lidar, xian2022location} extract foreground points according to semantic predictions and utilize heuristic clustering algorithms to assign instance IDs to thing points. These methods mainly focus on designing modules for predicting center regression or embedding vectors to enhance clustering. For example, DS-Net \cite{hong2021lidar, hong2024unified}, Panoptic-PolarNet \cite{zhou2021panoptic}, and Panoptic-PHNet \cite{li2022panoptic} improved clustering performance by designing a learnable clustering module and pseudo heatmap, among other techniques.
However, these methods typically handle semantic and instance segmentation separately, whereas our DQFormer proposes a unified workfolow for predicting both object and stuff classes.

\noindent\textbf{Center-based method} \cite{mei2023centerlps} utilizes object centers as queries to segment instances, eliminating detection or clustering processes. However, semantic segmentation remains indispensable for retrieving object centers in this method, while DQFormer proposes things queries directly from the BEV space.

\noindent\textbf{Query-based methods} \cite{gu2022maskrange, marcuzzi2023mask, su2023pups} utilize a set of learnable queries to implement unified LiDAR panoptic segmentation by predicting a set of non-overlapping binary masks and semantic classes for either a stuff class or a potential object. Specifically, MaskRange \cite{gu2022maskrange} and MaskPLS \cite{marcuzzi2023mask} employ learnable queries for range-based and point-based LiDAR panoptic segmentation, respectively. PUPS \cite{su2023pups} uses point-level classifiers to predict semantic masks and instance groups directly.
P3Former \cite{xiao2023position} introduces a mixed-parameterized positional embedding to guide the mask prediction and query update processes iteratively. However, all of these methods ignore the distinction between things and stuff in LiDAR scenes and face the challenge of mutual competition.
In contrast, our proposed DQFormer adopts the query-based paradigm but decouples things/stuff queries based on their intrinsic characteristics. This enables separate decoding to alleviate mutual competition.

\subsection{Query-based 2D/3D Segmentation}
Following the success of DETR \cite{carion2020end, zong2023detrs} in 2D detection, query-based segmentation methods \cite{petrovai2022semantic, mei2021transvos} have emerged as effective approaches for improving segmentation accuracy and efficiency, such as Panoptic-FCN \cite{li2021fully}, K-Net \cite{zhang2021k}, MaskFormer \cite{cheng2021per}, Mask2Former \cite{cheng2022masked}, and Panoptic SegFormer \cite{li2022panoptic}. These methods utilize queries to guide the segmentation process and employ various techniques like kernel updates, learnable queries with transformers, masked attention,~\etc.

Building on the impressive achievements of these methods in 2D domains, some approaches have extended the query-based paradigm to 3D segmentation~\cite{liu2024multi, athar20234d}. DyCo3D \cite{he2021dyco3d, he2022dynamic} and DKNet \cite{wu20223d} represent 3D instances using 1D kernels for decoding instance masks. CenterLPS \cite{mei2023centerlps} localizes and proposes instance queries by object centers for 3D instance and panoptic segmentation. Mask4D \cite{marcuzzi2023mask4d} extends the mask-based 3D panoptic segmentation model to 4D by reusing queries that decoded instances in previous scans. Our DQFormer maintains the intrinsic properties of point clouds while following a unified query-based workflow.

\section{Method}
\subsection{Overview}
As illustrated in Figure \ref{fig:pipeline}, DQFormer consists of three key modules: a multi-scale feature encoder, a decoupled query generator, and a query-oriented mask decoder. Specifically, (1) The feature encoder (Sec. \ref{Sec.FE}) extracts voxel-level features and point-wise embeddings at multiple resolutions. (2) The query generator (Sec. \ref{Sec.QG}) is designed to produce informative things/stuff queries with assigned semantics according to their positions and embeddings in BEV space. (3) The mask decoder (Sec. \ref{Sec.MD}) decodes segmentation masks by performing masked cross-attention between queries and point embeddings. Finally, the decoded masks are combined with the semantics of queries to produce the panoptic results.
In this section, we elaborate on the above components as well as the training scheme.

\subsection{Multi-scale Feature Encoder} \label{Sec.FE}
Firstly, we introduce a sparse voxel-based backbone to encode input point clouds, extracting multi-scale voxel features for the \emph{query generator} and point-wise embeddings for the \emph{mask decoder}.

Specifically, given an input point cloud $P \in \mathbb{R}^{N_p\times4}$ (coordinates and intensity), we perform 3D grid voxelization to obtain voxel-point indices. Then, each voxel feature is extracted by feeding the representations $f_p$ of points within the same voxel into MLPs and applying max-pooling across the point dimension. Here, the point representations $f_p$ consist of concatenating the point coordinates, intensity, and the offset of the point-to-voxel center. Consequently, we form sparse voxel features denoted as $F^v \in \mathbb{R}^{N_v \times 32}$ with a dense spatial resolution of $H \times W \times D$, where $N_v$ is the number of sparse voxels, $H$, $W$, and $D$ represent the length, width, and height of the voxelized space, respectively.

Furthermore, we implement a UNet-like architecture to extract voxel features of multi-resolutions. Each resolution $level_i$ comprises an encoder to aggregate long-range information using the radial window self-attention proposed in \cite{lai2023spherical}. It also includes a down-sampling module to extract sub-voxel features at lower resolutions, along with a decoder that up-samples the lower-resolution features and integrates them with the encoded features to generate voxel features at resolution $level_i$. In practice, we implement a four-layer feature extractor to encode multi-scale voxel features $(F_1^v, F_2^v, F_3^v, F_4^v)$. These features are further interpolated to points of full resolution using a $k$-nearest-neighbor weighted summation, bringing them to the resolution of the original point cloud. This operation is denoted as the V2P (Voxel-to-Point) operation. As a result, it also outputs multi-scale point-wise embeddings $(F_1^p, F_2^p, F_3^p, F_4^p)$, where each embedding is interpolated from voxel features at resolution $level_i$, implying multi-scale contextual and geometric information.

\subsection{Decoupled Query Generator} \label{Sec.QG}
Due to the sparsity of point clouds, generating informative queries for decoding corresponding segmentation masks is crucial. In this work, we generate query proposals that encapsulate the features of instances/stuff based on their positions and embeddings in the Bird's Eye View (BEV) space.

\textbf{Decoupled Query Proposal.} 
To explicitly localize small instances while enlarging the receptive field for stuff, we propose a query proposal network to generate things/stuff queries within BEV embeddings of varying resolutions. 

\subsubsection{BEV Embedding Extraction} Initially, we project the voxel feature along the z-axis to generate the BEV feature, denoted as the V2B (Voxel-to-BEV) operation. Specifically, given the voxel feature $F_i^v$ at $level_i$ with dense spatial resolution $H_i \times W_i \times D_i$, we first concatenate the height dimension $D_i$ with the feature dimension $C_i$, and then adopt stacks of 2D CNN with channel-wise and spatial attention to encode the BEV embedding, denoted as $F_i^{bev} \in \mathbb{R}^{C_e \times H_i \times W_i}$, where $C_e$ represents the feature dimension in the embedding space. This BEV embedding serves as the shared feature map used to locate and classify each instance or stuff class.

\subsubsection{BEV Heatmap Prediction} Following \cite{li2021fully}, we utilize \emph{object centers} to indicate the positions of individual instances and \emph{stuff regions} to represent specific stuff categories. This implies that stuff regions sharing the same semantics are collectively considered as a single instance. Concretely, an object center head, composed of a series of 2D convolutions, is introduced to predict the center heatmap $M^{th}_{i} \in \mathbb{R} ^ {N{th} \times H_i \times W_i}$ at $level_i$. Here, $N_{th}$ represents the number of semantic categories for instances, and each channel of the heatmap represents the potential centers for one class. Meanwhile, a stuff region head, constructed using shallow 2D transformer decoder layers \cite{li2022panoptic}, is responsible for predicting the stuff region map $M^{st}_{i} \in \mathbb{R} ^ {N{st} \times H_i \times W_i}$, where $N_{st}$ denotes the number of semantic categories for stuff classes.

\begin{figure}[t]
\centering
    \includegraphics[width=0.48\textwidth]{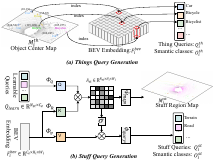}
    \vspace{-1.5em}
    \caption{\textbf{Details of query proposal generation:} Things queries are extracted from the BEV embedding at the corresponding positions. Stuff queries are generated using the learnable-query approach within the BEV space.}
    \vspace{-1em}
    \label{fig:query}
\end{figure}

\subsubsection{Query Proposals Generation} Next, we generate informative things/stuff queries based on heatmaps $M^{th}_{i}$ and $M^{st}_{i}$, along with the BEV embedding $F_i^{bev}$. 

For things queries generation, as shown in Figure \ref{fig:query}(a), we first select the top-$N_q$ positions on $M^{th}_{i}$ with the highest predicted scores, indicating the locations of potential instances. Then, we extract the embeddings from $F_i^{bev}$ at corresponding positions to represent the query weight for each selected instance position. For illustration, assuming a candidate position $(x_c,y_c)$ located in the $c$-channel on $M^{th}_i$, the embedding $F_i^{bev}[:,x_c,y_c] \in \mathbb{R} ^ {C_e \times 1 \times 1}$ is extracted to represent the query weight for this instance with predicted category $c$. Since the center heatmap indicates potential instance locations, the BEV embeddings at these specific positions contain more informative features of the existing instances. Consequently, we can obtain the things query proposals at $level_i$, denoted as $Q_i^{th}\in\mathbb{R}^{N_q \times C_e}$ with predicted semantic categories $O_i^{th}$, where $N_q$ is the number of instance queries in a scan that we set empirically.

During stuff queries generation, as depicted in Figure \ref{fig:query}(b), since stuff points are widely distributed across the entire scene, it is crucial to incorporate global context to capture more meaningful features. To achieve this, we utilize class-fixed learnable queries denoted as $Q^{learn}\in \mathbb{R}^{N_{st} \times C_e}$ for cross-attention with BEV features, where $N_{st}$ is the number of stuff classes. This ensures that each query aligns with the relevant BEV embeddings by predicting stuff regions from attention maps, formulated as follows:

\begin{equation}
    A^{st} = \frac{ \phi_{q}(Q^{learn}) \cdot \phi_{k}(F_i^{bev})}{\sqrt{C_e}}
\end{equation}
\begin{equation}
    M_i^{st} = \phi_{map}(A^{st}), Q_i^{st} = \phi_{query}[\sigma(A^{st}) \cdot \phi_{v}(F_i^{bev})]
\end{equation}
where $\phi_{*}$ represents linear layers, $\sigma$ denotes the softmax function, $A^{st} \in \mathbb{R} ^ {N_{st} \times H_iW_i}$ are the attention maps, and $Q_i^{st}\in \mathbb{R}^{N_{st} \times C_e}$ represents the stuff query proposals at $level_i$, associated with corresponding semantic categories $O_i^{st}$. This formulation effectively enlarges the receptive fields of stuff queries, allowing them to capture more information throughout the scene while incurring limited computational overhead within the BEV.

\textbf{Decoupled Query Fusion.} 
With the queries generated by the multi-scale query proposal network, we further design a query fusion module to merge the query proposals from different resolutions effectively.

\emph{1) Things Query Proposals Fusion:} Our objective is to merge queries at similar positions from multi-scale BEV embeddings, thereby enhancing the query representation for individual instances. To ensure intra-semantic consistency, we only fuse queries that share the same semantics. Specifically, we employ average pooling to fuse queries whose positions are within the same small window in the BEV, and their cosine similarities exceed a given threshold, denoted as $\theta_{th}$. This approach ensures that the window constrains the geometric consistency of queries while cosine similarity maintains feature consistency in the embedding space. With this proposed approach, we fuse the multi-scale query proposals into an integral set of things queries, denoted as $Q^{th} \in \mathbb{R} ^{N_t \times C_e}$, along with predicted semantic categories $O^{th} \in \mathbb{R} ^ {N_t}$, where $N_t$ represents the predicted number of objects in the current scan.

\emph{2) Stuff Query Proposals Fusion:} We aim to merge queries with the same semantics to integrate multi-scale global context for each stuff class. Initially, we determine the presence of each stuff class based on the stuff region maps. Specifically, the presence of pixels with values surpassing a given threshold, denoted as $\theta_{st}$, on the stuff region maps indicates the existence of corresponding stuff points within the scene. Subsequently, we select the existing stuff queries and employ average summation to fuse queries with the same semantic categories. This process integrates queries from multiple scales that share the same semantics into a single embedding. It enhances each stuff query with global receptive fields while maintaining semantic consistency, resulting in the set of stuff queries denoted as $Q^{st} \in \mathbb{R}^{N_s \times C_e}$ with associated semantic categories $O^{st} \in \mathbb{R}^{N_s}$, where $N_s$ represents the number of existing stuff classes.

\begin{figure}[t]
\centering
    \includegraphics[width=0.46\textwidth]{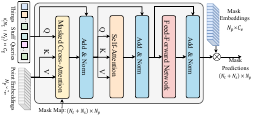}
    \vspace{-1.2em}
    \caption{\textbf{Detailed pipeline of the decoder block:} consisting of masked cross-attention, self-attention, and a feed-forward network.}
    \vspace{-1.3em}
    \label{fig:decoder}
\end{figure}

\subsection{Query-oriented Mask Decoder} \label{Sec.MD}
With the decoupled queries for objects and background encapsulating informative features, we introduce a query-oriented mask decoder to parse segmentation masks. It performs multi-level masked cross-attention guided by these queries. Since both types of queries, assigned with semantics, share a consistent format, we can decode their corresponding segmentation masks within a unified workflow.

As depicted in Figure \ref{fig:decoder}, our mask decoder comprises multiple decoder blocks, each consisting of masked cross-attention at a specific resolution, followed by self-attention and a feed-forward network (FFN). Specifically, in the decoder block of $level_i$, we first perform masked cross-attention between the concatenated queries $Q \in \mathbb{R}^{(N_t+N_s) \times C_e}$ and point embeddings $F_i^p \in \mathbb{R}^{N_p \times C_e}$. The mask map $M_i \in \mathbb{R}^{(N_t+N_s) \times Np}$, indicating the noteworthy key points, is generated from the previous block. Then, a self-attention layer is utilized to establish context between queries, and the FFN is employed to enhance the query representations. Finally, the segmentation mask is generated via the dot product between the output queries and point-wise mask embeddings $E \in \mathbb{R}^{N_p \times C_e}$, along with the sigmoid activation function. The mask decoding process is expressed as follows:
\begin{equation}
    Q' = \sigma[\frac{\phi_q(Q) \cdot \phi_k(F^p_i)^T}{\sqrt{C_e}} \odot M_{i-1}]\cdot \phi_v(F^p_i) + Q
\end{equation}
\begin{equation}
    Q'' = \mathrm{FFN}[\sigma(\frac{\varphi_q(Q') \cdot \varphi_k(Q')^T}{\sqrt{C_e}}) \cdot \varphi_v(Q') + Q']
\end{equation}
\begin{equation}
    M_i = \text{Sigmoid}( Q'' \cdot E^T) 
\end{equation}
where $\phi_*$ and $\varphi_*$ represent linear layers, $\odot$ denotes the Hadamard product, and $\sigma$ represents the softmax function. For simplicity, we omit LayerNorm in the formula. It's worth noting that the mask embedding $E$ is composed of the summation of full resolution point features $F_4^p$ and point-wise positional encoding \cite{vaswani2017attention} $P_e \in \mathbb{R}^{N_p \times C_e}$, defined as $E = F_4^p + P_e$.

During training, deep supervision is applied to the multi-level mask predictions ${M_1, M_2, ..., M_L}$. In the inference phase, we utilize the masks from the last decoder block and apply the mask fusion module \cite{mei2023centerlps} to integrate duplicate masks. Subsequently, these binary masks are combined with the semantics of queries $\{O^{th}, O^{st}\}$ to generate the panoptic results.

\subsection{Loss Function}
In training, we supervise the \emph{object center} and \emph{stuff region} heatmaps for the localization and classification of objects and background regions:
\begin{equation}
    \mathcal{L}_{hm} = \sum_{i} \text{FL}(M_i^{th}, Y_i^{th}) / N_q + \sum_{i} \text{FL}(M_i^{st}, Y_i^{st})/H_iW_i
\end{equation}
where $\text{FL}(\cdot,\cdot)$ denotes the Focal Loss\cite{lin2017focal}, $Y_i^{th}$ and $Y_i^{st}$ represent the ground truth of $M_i^{th}$ and $M_i^{st}$, respectively. Following \cite{li2021fully, zhou2022centerformer}, we assign the center of an instance with semantic category $c$ to the $c$-th channel of $Y_i^{th}$ using a Gaussian kernel. $Y_i^{st}$ is generated by interpolating the one-hot semantic label in the BEV space to corresponding sizes.

Meanwhile, we also supervise the mask predictions for segmentation using binary cross-entropy and dice loss:
\begin{equation}
    \mathcal{L}_{mask} = \sum_{i} \text{BCE}(M_i, Y) + \sum_{i} \text{Dice}(M_i, Y)
\end{equation}
where $Y$ represents the ground truth masks matched with predictions. Specifically, instance masks are matched with ground truth through the BEV positions, while stuff masks are matched in a one-to-one manner.

To enhance the point-wise embeddings, we also add an auxiliary MLP head to $F_4^p$ and employ a semantic loss $\mathcal{L}_{sem}$ to guide the class distribution of points. Overall, DQFormer can be trained end-to-end with the above loss:
\begin{equation}
    \mathcal{L} = \mathcal{L}_{hm} + \mathcal{L}_{mask} + \mathcal{L}_{sem}
\end{equation}

\section{Experiments}
In this section, we introduce the implementation details and present comprehensive experiments conducted on the nuScenes \cite{caesar2020nuscenes} and SemanticKITTI \cite{behley2019semantickitti} datasets to showcase the state-of-the-art performance achieved by DQFormer in large-scale outdoor scenarios. Furthermore, we conduct ablation studies on key components of our model to validate the efficacy and impact of the proposed modules. Additionally, visualizations and qualitative analyses are provided to underscore the efficacy of our model. 

\subsection{Datasets and Metrics}
\textbf{nuScenes} \cite{caesar2020nuscenes} dataset is a comprehensive urban driving dataset comprising 1000 LiDAR scenes, each spanning a duration of 20 seconds, captured using a 32-beam LiDAR sensor. This extensive dataset is divided into 850 scenes allocated for training and validation, with an additional 150 scenes for testing. For the LiDAR panoptic segmentation task, nuScenes annotates 16 distinct point-wise labels, including 10 \emph{thing} categories and 6 \emph{stuff} categories.

\textbf{SemanticKITTI} \cite{behley2019semantickitti} is based on KITTI \cite{geiger2012we} odometry dataset, which collects 22 LiDAR sequences with a Velodyne HDL-64 laser scanner. Among these sequences, 10 are allocated for training, 1 for validation, and 10 for testing. The dataset includes 19 annotated point-wise labels for the LiDAR-based panoptic segmentation task, consisting of 8 \emph{thing} classes and 11 \emph{stuff} classes.

\textbf{Metrics.} The metrics \cite{behley2021benchmark} for LiDAR-based panoptic segmentation include Panoptic Quality (PQ), Segmentation Quality (SQ), and Recognition Quality (RQ), which are formulated as:
\begin{equation}
    \vspace{-5pt}
    \mathrm{PQ} = \underbrace{\frac{\sum_{\textbf{TP}}\textbf{IoU}}{|\textbf{TP}|}}_\text{SQ} \times \underbrace{\frac{|\textbf{TP}|}{|\textbf{TP}|+\frac{1}{2}|\textbf{TN}|+\frac{1}{2}|\textbf{FP}|}}_\text{RQ}.
\end{equation}
These metrics are also calculated separately for \emph{thing} and \emph{stuff} classes indicated by PQ$^\mathrm{Th}$, SQ$^\mathrm{Th}$, RQ$^\mathrm{Th}$ and PQ$^\mathrm{St}$, SQ$^\mathrm{St}$, RQ$^\mathrm{St}$. In addition, we also report PQ$^\mathrm{\dag}$, as defined in \cite{porzi2019seamless}, which utilizes SQ as PQ for \emph{stuff} classes.

\subsection{Implementation Details}
For the voxelization process, we discretize the 3D space within $[[\pm51.2,m], [\pm51.2,m], [-4,m, 2.4,m]]$ into voxels with a voxelization resolution of $[0.05,m, 0.05,m, 0.05,m]$. Multi-scale voxel features $F_i^v$ are obtained at resolutions corresponding to $[\frac{1}{8}, \frac{1}{4}, \frac{1}{2}, 1]$ of the original dense resolutions. The number of semantic categories for both \emph{things}, $N_{th}$, and \emph{stuff}, $N_{st}$, are determined based on the annotations of datasets. We set the number of instance queries in a scan, $N_q$, to 150. We use the threshold $\theta_{th}=0.85$ to discriminate cosine similarities between \emph{things} queries and the threshold $\theta_{st}=0.5$ to indicate the existence of \emph{stuff} regions. The mask decoder consists of $N_l=3$ decoder blocks, each block employing point embeddings interpolated from voxel features with resolutions of $[\frac{1}{8}, \frac{1}{4}, \frac{1}{2}]$, respectively. During training, we adopt global rotation, global scaling, and random flip for data augmentations. We train the models for 80 epochs using the AdamW optimizer \cite{loshchilov2017decoupled} on 8 NVIDIA RTX A6000 GPUs. The initial learning rate is set to $1e$-4 and is decayed by a factor of 10 at epoch 60.

\begin{table*}[t]
    \caption{\textbf{Comparison of LiDAR panoptic segmentation performance on the test set of nuScenes}. All results in [\%].}
    \vspace{-.5em}
    \centering
    \definecolor{mycolor}{rgb}{0.64, 0.87, 0.93}
    \small
    \setlength{\tabcolsep}{0.01\linewidth}
    \scalebox{0.88}{
    \begin{tabular}{c|c|>{\columncolor{mycolor!25}}c|ccc|ccc|ccc} 
        \toprule
        Type & Method & PQ & PQ$^\mathrm{\dag}$ & RQ& SQ & PQ$^\mathrm{Th}$ &RQ$^\mathrm{Th}$ & SQ$^\mathrm{Th}$ & PQ$^\mathrm{St}$ &RQ$^\mathrm{St}$ & $\mathrm{SQ^{St}}$ \\
        \midrule
        \multirow{6}{*}{\makecell{Detection-\\ based}} & PanopticTrackNet \cite{hurtado2020mopt} & 51.6 & 56.1 & 63.3 & 80.4 & 45.9 & 56.1 & 81.4 & 61.0 & 75.4 & 79.0 \\
        & EfficientLPS \cite{sirohi2021efficientlps} & 62.4 & 66.0 & 74.1 & 83.7 & 57.2 & 68.2 & 83.6 & 71.1 & 84.0 & 83.8\\
        & AOP-Net \cite{xu2023aop} & 68.3 & - & 78.2 & 86.9 & 67.3 & 75.6 & 88.6 & 69.8 & 82.6 & 84.0\\
        & SPVNAS \cite{tang2020searching} + CenterPoint \cite{yin2021center} & 72.2 & 76.0 & 81.2 & 88.5 & 71.7 & 79.4 & 89.7 & 73.2 & 84.2 & 86.4\\
        & Cylinder3D++ \cite{zhu2021cylindrical} + CenterPoint \cite{yin2021center} & 76.5 & 79.4 & 85.0 & 89.6 & 76.8 & 84.0 & 91.1 & 76.0 & 86.6 & 87.2\\
        & (AF)2-S3Net \cite{cheng20212} + CenterPoint \cite{yin2021center} & 76.8 & 80.6 & 85.4 & 89.5 & 79.8 & 86.8 & 91.8 & 71.8 & 83.0 & 85.7\\
        \midrule
        \multirow{4}{*}{\makecell{Clustering-\\ based}} & Panoptic-PolarNet \cite{zhou2021panoptic} & 63.6 & 67.1 & 75.1 & 84.3 & 59.0 & 69.8 & 84.3 & 71.3 & 83.9 & 84.2 \\
        & PolarStream \cite{chen2021polarstream} & 70.9 & 74.4 & 81.7 & 85.9 & 70.3 & 80.3 & 86.7 & 71.7 & 84.2 & 84.4 \\
        & LCPS(LiDAR) \cite{zhang2023lidar} & 72.8 & 76.3 & 81.7 & 88.6 & 72.4 & 80.0 & 90.2 & 73.5 & 84.6 & 86.1 \\
        & CPSeg \cite{li2023cpseg} & 73.2 & 76.3 & \textbf{82.7} & 88.1 & 72.9 & \textbf{81.3} & 89.2 & 74.0 & \textbf{85.0} & \textbf{86.3} \\
        \midrule
        \multirow{2}{*}{Query-based} & MaskPLS \cite{marcuzzi2023mask} & 61.1 & 64.3 & 68.5 & 86.8 & 54.3 & 58.8 & 87.8 & 72.4 & 63.4 & 85.1 \\
        & \cellcolor{gray!10}DQFormer (Ours) & \cellcolor{mycolor!25}\textbf{73.9} & \cellcolor{gray!10}\textbf{76.8} & \cellcolor{gray!10}82.4 & \cellcolor{gray!10}\textbf{89.6} & \cellcolor{gray!10}\textbf{74.4} & \cellcolor{gray!10}80.7 & \cellcolor{gray!10}\textbf{91.9} & \cellcolor{gray!10}\textbf{78.2} & \cellcolor{gray!10}\textbf{85.0} & \cellcolor{gray!10}85.8 \\
        \bottomrule
    \end{tabular}}
    \vspace{-.5em}
    \label{tab:nuscenes_test}
\end{table*}

\begin{table*}[t]
    \caption{\textbf{Comparison of LiDAR panoptic segmentation performance on the validation set of nuScenes}. All results in [\%].}
    \vspace{-.5em}
    \centering
    \definecolor{mycolor}{rgb}{0.64, 0.87, 0.93}
    \small
    \setlength{\tabcolsep}{0.012\linewidth}
    \scalebox{0.88}{
    \begin{tabular}{c|c|>{\columncolor{mycolor!25}}c|ccc|ccc|ccc} \toprule
        Type & Method & PQ & PQ$^\mathrm{\dag}$ & RQ& SQ & PQ$^\mathrm{Th}$ &RQ$^\mathrm{Th}$ & SQ$^\mathrm{Th}$ & PQ$^\mathrm{St}$ &RQ$^\mathrm{St}$ & $\mathrm{SQ^{St}}$ \\
        \midrule
        \multirow{2}{*}{\makecell{Detection-\\ based}} & PanopticTrackNet \cite{hurtado2020mopt} & 51.4 & 56.2 & 63.3 & 80.2 & 45.8 & 55.9 & 81.4 & 60.4 & 75.5 & 78.3 \\
        & EfficientLPS \cite{sirohi2021efficientlps} & 62.0 & 65.6 & 73.9 & 83.4 & 56.8 & 68.0 & 83.2 & 70.6 & 83.6 & 83.8\\
        \midrule
        
        \multirow{6}{*}{\makecell{Clustering-\\ based}} & DS-Net \cite{hong2021lidar} & 42.5 & 51.0 & 50.3 & 83.6 & 32.5 & 38.3 & 83.1 & 59.2 & 70.3 & 84.4 \\
        & GP-S3Net \cite{razani2021gp} & 61.0 & 67.5 & 72.0 & 84.1 & 56.0 & 65.2 & 85.3 & 66.0 & 78.7 & 82.9 \\
        & Panoptic-PolarNet \cite{zhou2021panoptic} & 63.4 & 67.2 & 75.3 & 83.9 & 59.2 & 70.3 & 84.1 & 70.4 & 83.5 & 83.6 \\
        & PVCL \cite{liu2022prototype} & 64.9 & 67.8 & 77.9 & 81.6 & 59.2 & 72.5 & 79.7 & 67.6 & 79.1 & 77.3 \\
        & SCAN \cite{xu2022sparse} & 65.1 & 68.9 & 75.3 & 85.7 & 60.6 & 70.2 & 85.7 & 72.5 & 83.8 & 85.7 \\
        & Panoptic-PHNet \cite{li2022PHNet} & 74.7 & 77.7 & 84.2 & \textbf{88.2} & 74.0 & 82.5 & 89.0 & 75.9 & 86.9 & 86.8 \\
        \midrule
        
        \multirow{4}{*}{\makecell{Query-\\ based}} & MaskPLS-M \cite{marcuzzi2023mask} & 57.7 & 60.2 & 66.0 & 71.8 & 64.4 & 73.3 & 84.8 & 52.2 & 60.7 & 62.4\\
        & PUPS \cite{su2023pups} & 74.7 & 77.3 & 83.3 & 89.4 & 75.4 & 81.9 & 91.8 & 73.6 & 85.6 & 85.6 \\
        & P3Former \cite{xiao2023position} & 75.9 & 78.9 & 84.7 & 89.7 & 76.8 & 83.3 & \textbf{92.0} & 75.4 & 87.1 & 86.0 \\
        & \cellcolor{gray!10}DQFormer (Ours) & \cellcolor{mycolor!25}\textbf{77.7} & \cellcolor{gray!10}\textbf{79.5} & \cellcolor{gray!10}\textbf{89.2} & \cellcolor{gray!10}86.8 & \cellcolor{gray!10}\textbf{77.8} & \cellcolor{gray!10}\textbf{89.5} & \cellcolor{gray!10}86.7 & \cellcolor{gray!10}\textbf{77.5} & \cellcolor{gray!10}\textbf{88.6} & \cellcolor{gray!10}\textbf{87.0} \\ \bottomrule
    \end{tabular}}
    \vspace{-1em}
    \label{tab:nuscenes_val}
\end{table*}

\begin{table}[t]
    \caption{\textbf{Comparison of LiDAR panoptic segmentation performance on the test set of SemanticKITTI} \cite{behley2019semantickitti}. All results in [\%].}
    \vspace{-.5em}
    \centering
    \definecolor{mycolor}{rgb}{0.64, 0.87, 0.93}
    \small
    \setlength{\tabcolsep}{0.009\linewidth}
    \scalebox{0.88}{
    \begin{tabular}{c|c|>{\columncolor{mycolor!25}}c|ccc} \toprule
        Type & Method & PQ & PQ$^\mathrm{\dag}$ & RQ& SQ \\
        \midrule
        \multirow{3}{*}{\makecell{Detection-\\ based}} & RangeNet++\cite{milioto2019rangenet++} + PointPillars\cite{lang2019pointpillars} & 37.1 & 45.9 & 47.0 & 75.9  \\
        & KPConv\cite{thomas2019kpconv} + PointPillars\cite{lang2019pointpillars} & 44.5 & 52.5 & 54.4 & 80.0  \\
        & EfficientLPS\cite{sirohi2021efficientlps} & 57.4 & 63.2 & 68.7 & 83.0  \\
        \midrule
        
        \multirow{7}{*}{\makecell{Clustering-\\ based}} & LPSAD\cite{milioto2020lidar} & 38.0 & 47.0 & 48.2 & 76.5  \\
        & Panoster\cite{gasperini2021panoster} & 52.7 & 59.9 & 64.1 & 80.7  \\
        & Panoptic-PolarNet\cite{zhou2021panoptic} & 54.1 & 60.7 & 65.0 & 81.4 \\
        & DS-Net\cite{hong2021lidar} & 55.9 & 62.5 & 66.7 & 82.3 \\
        & CPSeg \cite{li2021cpseg} & 57.0 & 63.5 & 68.8 & 82.2 \\
        & SCAN \cite{xu2022sparse} & 61.5 & 67.5 & 72.1 & 84.5 \\
        & Panoptic-PHNet \cite{li2022PHNet} & 61.5 & 67.9 & 72.1 & 84.8  \\
        \midrule

        \multirow{1}{*}{\makecell{Center-based}} & CenterLPS\cite{mei2023centerlps} & 61.6 & 67.9 & 72.6 & 84.0  \\
        \midrule
        
        \multirow{3}{*}{\makecell{Query-\\ based}} &  MaskPLS-M\cite{marcuzzi2023mask} & 58.2 & 63.3 & 68.6 & 83.9 \\
        & PUPS\cite{su2023pups} & 62.2 & 65.8 & 72.8 & 84.2 \\
        & \cellcolor{gray!10}DQFormer (ours) & \cellcolor{mycolor!25}\textbf{63.1} & \cellcolor{gray!10}\textbf{67.9} & \cellcolor{gray!10}\textbf{73.6} & \cellcolor{gray!10}\textbf{85.0} \\ 
        % & \rowcolor{gray!10} DQFormer$^\dag$ (ours) & \cellcolor{mycolor!25}\textbf{64.9} & \textbf{69.5} & \textbf{75.1} & \textbf{85.8} \\
        \bottomrule
    \end{tabular}}
    \vspace{-.5em}
    \label{tab:semkitti_test}
\end{table}

\begin{table}[t]
    % \vspace{-0.2cm}
    \caption{\textbf{Comparison of LiDAR panoptic segmentation performance on the validation set of SemanticKITTI}\cite{behley2019semantickitti}. Bold and underlined indicate the best and second-best performances, respectively. All results in [\%].}
    \vspace{-.5em}
    \centering
    \definecolor{mycolor}{rgb}{0.67, 0.9, 0.93}
    \small
    \setlength{\tabcolsep}{0.009\linewidth}
    \scalebox{0.88}{
    \begin{tabular}{c|c|>{\columncolor{mycolor!25}}c|ccc} \toprule
        Type & Method & PQ & PQ$^\mathrm{\dag}$ & RQ & SQ \\
        \midrule
        \multirow{4}{*}{\makecell{Detection-\\ based}} &  RangeNet++\cite{milioto2019rangenet++} + PointPillars\cite{lang2019pointpillars} & 36.5 & - & 44.9 & 73.0 \\
        & PanopticTrackNet\cite{hurtado2020mopt} & 40.0 & - & 48.3 & 73.0 \\
        & KPConv\cite{thomas2019kpconv} + PointPillars\cite{lang2019pointpillars} & 41.1 & - & 50.3 & 74.3 \\
        & EfficientLPS\cite{sirohi2021efficientlps} & 59.2 & 65.1 & 69.8 & 75.0 \\
        \midrule

        \multirow{4}{*}{\makecell{Clustering-\\ based}} &  LPSAD\cite{milioto2020lidar} & 36.5 & 46.1 & - & -  \\
        & Panoster\cite{gasperini2021panoster} & 55.6 & - & 66.8 & 79.9  \\
        & DS-Net\cite{hong2021lidar} & 57.7 & 63.4 & 68.0 & 77.6  \\
        & Panoptic-PolarNet\cite{zhou2021panoptic} & 59.1 & 64.1 & 70.2 & 78.3 \\
        \midrule

        \multirow{1}{*}{\makecell{Center-based}} &  CenterLPS\cite{mei2023centerlps} & 62.1 & 67.0 & 72.0 & 80.7 \\
        \midrule
        
        \multirow{3}{*}{\makecell{Query-\\ based}} &  MaskPLS-M\cite{marcuzzi2023mask} & 59.8 & - & 69.0 & 76.3 \\
        & PUPS\cite{su2023pups} & \textbf{64.4} & \textbf{68.6} & \textbf{74.1} & \underline{81.5} \\
        & \cellcolor{gray!10}DQFormer (ours) & \cellcolor{mycolor!25}\underline{63.5} & \cellcolor{gray!10}\underline{67.2} & \cellcolor{gray!10}\underline{73.1} & \cellcolor{gray!10}\textbf{81.7} \\ \bottomrule
    \end{tabular}}
    \vspace{-.5em}
    \label{tab:semkitti_val}
\end{table}

\subsection{Comparison with the state-of-the-art.}
\textbf{Results on nuScenes.}
Table \ref{tab:nuscenes_test} and Table \ref{tab:nuscenes_val} present the comparison results between our DQFormer and other state-of-the-art methods on the nuScenes \cite{caesar2020nuscenes} test and validation sets.

\subsubsection{Compared with detection-based methods} 
DQFormer demonstrates a significant improvement against single-architecture methods, particularly outperforming EfficientLPS \cite{sirohi2021efficientlps} and AOPNet \cite{xu2023aop} by a large margin (+17.2\% and +7.4\%) in terms of PQ$^\mathrm{Th}$. This highlights the efficacy of our query generator in efficiently localizing and classifying instances within the BEV, eliminating the need for unnecessary bounding box predictions. 
Compared to methods that combine two architectures (\ie semantic segmentation models \cite{tang2020searching, zhu2021cylindrical, cheng20212} and a 3D detection method CenterPoint \cite{yin2021center}), DQFormer achieves comparable performance and surpasses them in terms of SQ$^\mathrm{Th}$ by +2.2\%, 0.8\%, and 0.1\%, respectively. While these methods achieve superior performances on RQ$^\mathrm{Th}$ compared to DQFormer and other models, we attribute this to CenterPoint's superior detection performance on small targets, resulting in a high recall rate (RQ$^\mathrm{Th}$). However, these two-stage architectures are inefficient, while DQFormer effectively integrates semantic and instance segmentation tasks in a unified workflow.

\subsubsection{Compared with clustering-based methods}
They \cite{milioto2020lidar,gasperini2021panoster,hong2021lidar,zhou2021panoptic,li2022PHNet} generally employ separate semantic/instance branches, while our DQFormer achieves superior results within a unified workflow based on the query-based segmentation paradigm. For instance, compared to the state-of-the-art clustering-based method Panoptic-PHNet \cite{li2022PHNet}, we enhance the PQ from 74.7\% to 77.7\% on the validation set, marking a substantial gain of 3.8\% in PQ$^\mathrm{Th}$.

\subsubsection{Compared with query-based methods}
While MaskPLS \cite{marcuzzi2023mask}, PUPS \cite{su2023pups}, and P3Former \cite{xiao2023position} leverage the learnable query approach, our DQFormer demonstrates remarkable performance, surpassing them by 20.0\%, 3.0\%, and 1.8\% in terms of PQ on the validation set, respectively. Compared to P3Former, which introduces a mixed-parameterized positional embedding for better distinguishing various instances, our DQFormer decouples queries and leads to gains on both things and stuff simultaneously (i.e., +1.0\% on PQ$^\mathrm{Th}$ and +2.1\% on PQ$^\mathrm{St}$). This substantiates the effectiveness of our query decoupling strategy, enabling the query embeddings to contain more informative features.
It is worth noting that while PUPS and P3Former have not released their performances on the test set, DQFormer still outperforms MaskPLS by 12.8\% on the test split.

\textbf{Results on SemanticKITTI.}
We validate the effectiveness and generalization of our method on the test and validation sets of SemanticKITTI \cite{behley2019semantickitti} as shown in Table \ref{tab:semkitti_test} and Table \ref{tab:semkitti_val}.

DQFormer surpasses all detection-based and clustering-based methods on both the test and validation splits, achieving 5.7\% higher PQ compared to EfficientLPS \cite{sirohi2021efficientlps} and 1.6\% higher PQ compared to Panoptic-PHNet \cite{li2022PHNet}, respectively. Notably, it outperforms CenterLPS \cite{mei2023centerlps}, a recent center-based method utilizing object queries for instance segmentation. On the contrary, DQFormer offers a more unified approach that employs things/stuff queries for panoptic segmentation, significantly improving PQ by 1.5\%.

Compared with query-based methods, DQFormer achieves notable gains in PQ compared to MaskPLS and PUPS on the test split, with improvements of 4.9\% and 0.9\%, respectively. While DQFormer performs slightly weaker than PUPS in terms of RQ on the validation split, this can be attributed to PUPS employing a CutMix strategy for data augmentations during training. In contrast, our model employs regular data augmentation techniques as previous methods to verify its generalization. Despite this, DQFormer still enhances the Segmentation Quality (SQ) by 0.8\% and 0.2\% on both the test and validation sets compared to PUPS. This underscores the effectiveness of DQFormer's spirit, which decouples things/stuff queries according to their intrinsic properties and disentangles classification/segmentation to mitigate ambiguity, promoting the segmentation quality.

\begin{table}[t]
    \caption{\textbf{Ablation on the network components} on SemanticKITTI validation set. QG denotes query generator. All scores are in [\%].}
    \vspace{-.5em}
    \centering
    \small
    \definecolor{mycolor}{rgb}{0.64, 0.87, 0.93}
    \resizebox{0.95\columnwidth}{!}{
    \setlength{\tabcolsep}{0.013\linewidth}
    \begin{tabular}{c|cccc|>{\columncolor{mycolor!25}}c|cccc} \toprule
        Variants & MD & ME & \makecell{QG \\ St. / Obj. / QF} & MF & PQ & RQ & SQ & PQ$^\mathrm{Th}$ & PQ$^\mathrm{St}$ \\
        \midrule
        0 & - & - & - \;  -  \;  -  & - & 60.6 & 70.4 & 76.0 & 63.1 & 58.8 \\
        1 & \cmark & - & - \;  -  \;  -  & - & 60.8 & 70.7 & 76.0 & 62.3 & 59.7 \\
        2 & \cmark & \cmark & - \;  -  \;  -  & - & 61.3 & 71.4 & 76.6 & 63.7 & 59.6 \\
        \midrule
        3 & \cmark & \cmark & \cmark \,\,  \xmark  \,\,  \xmark  & - & 61.7 & 71.5 & 76.4 & 64.2 & \textbf{59.9} \\
        4 & \cmark & \cmark & \cmark \,\,  \cmark  \,\,  \xmark  & - &  62.5 & 72.0 & 76.6 & 66.4 & 59.6  \\ % 175_v15
        5 & \cmark & \cmark & \cmark \,\,  \cmark  \,\,  \cmark  & - & 63.1 & 72.9 & 81.5 & 67.8 & 59.6 \\
        \midrule
        6 & \xmark & \cmark & \cmark \,\,  \cmark  \,\,  \cmark  & - &  62.1 & 71.5 & 76.5 & 65.8 & 59.4 \\
        7 & \cmark & \cmark & \cmark \,\,  \cmark  \,\,  \cmark  & \cmark & \textbf{63.5} & \textbf{73.1} & \textbf{81.7} & \textbf{68.8} & 59.6 \\
        \bottomrule
    \end{tabular}}
    \label{ab:1}
    \vspace{-0.5em}
\end{table}

\subsection{Ablation Study}
We conduct ablation studies on key components of our model, including network components, decoupling strategy, fusion modules, mask loss functions, and model settings with efficiency.

\textbf{Effects of Network Components.} 
Table \ref{ab:1} presents the ablation results of our proposed components. 

\subsubsection{Clustering-based baseline} We establish a clustering-based baseline in Variant 0 by equipping our backbone with the dynamic shift (DS) module \cite{hong2021lidar} to perform instance segmentation. For a fair comparison, it maintains the same semantic predictions as the backbone output of our DQFormer and achieves a Panoptic Quality (PQ) score of 60.6\%.

\subsubsection{Query-based baseline} Variant 1 and 2 serve as query-based baseline models that employ the query-based approach with vanilla mask decoder (MD). Variant 1 performs cross-attention between learnable queries and voxel features to produce segmentation results. This variant improves PQ$^\mathrm{St}$ by 0.9\% compared to Variant 0, highlighting the effectiveness of the query-based approach in enhancing the segmentation quality of stuff classes by introducing global context in the mask decoder. Variant 2 utilizes mask embedding (ME) instead of voxel features and exhibits a 1.4\% PQ$^\mathrm{Th}$ improvement over Variant 1. This improvement is attributed to the utilization of mask embeddings that are composed of point-wise features and positional encoding, which are more informative and essential for capturing small objects effectively.

\subsubsection{Effects of the query generator} Variants 3 and 4, which incorporate Stuff Region (St.) and Object Center (Obj.) heads, respectively, achieve PQ gains of 0.4\% and 1.2\% over Variant 2. Moreover, Variant 3 enhances PQ$^\mathrm{St}$ by 0.5\% while Variant 4 improves PQ$^\mathrm{Th}$ by 2.7\%. These improvements highlight the effectiveness of the query generator that ensures the query proposals are spatially close to the ground truth and contain practical information, contributing to improved segmentation performance. Variant 5 includes the Query Fusion Module (QF) and further enhances PQ by 0.6\% compared to Variant 4. This underscores the efficacy of multi-scale query fusion in integrating more comprehensive query embeddings, ultimately improving segmentation quality.

\subsubsection{Effects of the mask decoder} Variant 6 replaces the query-oriented mask decoder (MD) with a single-layer transformer decoder, resulting in a 5\% reduction in SQ compared to Variant 5. This reduction underscores the potency of the multi-level masked cross-attention mechanism present in the query-oriented mask decoder, which allows queries to focus on informative keypoints across different resolutions and contributes to the overall segmentation quality.

\subsubsection{Effects of the mask fusion} Variant 7 utilizes the mask fusion module (MF) \cite{mei2023centerlps} to integrate duplicated instance masks, leading to a 1.0\% improvement in PQ$^\mathrm{Th}$ compared to Variant 5. This module addresses the persistence of duplicated queries, particularly for large objects, by merging masks associated with the same object. This process ensures a more comprehensive segmentation by eliminating redundancy.

\begin{table}[t]
    \caption{\textbf{Ablation on the decoupling strategy} on SemanticKITTI validation set, with DQ (Decoupling Queries), HM (Hungarian Matching), and DC cls.\&seg. (Decoupling classification/segmentation). We don't use the mask fusion for a fair comparison. All scores are in [\%].}
    \vspace{-.5em}
    \centering
    \small
    \definecolor{mycolor}{rgb}{0.64, 0.87, 0.93}
    \resizebox{0.95\columnwidth}{!}{
    \setlength{\tabcolsep}{0.01\linewidth}
    \begin{tabular}{c|ccc|>{\columncolor{mycolor!25}}c|cccc} \toprule
        Variants & DQ & \makecell{Match \\ HM / BEV} & \makecell{DC cls.\&seg. \\ Thing / Stuff} &  PQ & RQ & SQ & PQ$^\mathrm{Th}$ & PQ$^\mathrm{St}$ \\
        \midrule
        Baseline & \xmark & \cmark \,\,\, - & - \,\,\,  - & 61.3 & 71.4 & 76.6 & 63.7 & 59.6 \\
        1 & \cmark & \cmark \,\,\, - & - \,\,\,  - & 61.4 & 71.5 & 76.6 & 64.0 & 59.6 \\
        2 & \cmark & \cmark \,\,\, - & \xmark \,\,\,  \cmark & 61.7 & 71.5 & 76.4 & 64.2 & \textbf{59.9} \\
        3 & \cmark & - \,\,\, \cmark  & \cmark \,\,\,  \xmark & 62.4 & 72.0 & 76.6 & 66.5 & 59.5 \\
        4 & \cmark & - \,\,\, \cmark & \cmark \,\,\,  \cmark & \textbf{63.1} & \textbf{72.9} & \textbf{81.5} & \textbf{67.8} & 59.6 \\
        \bottomrule
    \end{tabular}}
    \label{ab:2}
    \vspace{-1.0em}
\end{table}

\begin{table}[t]
    \caption{\textbf{Per-class PQ results of small instances} on the SemanticKITTI validation set. All scores are in [\%].}
    \centering
    \small
    \resizebox{0.95\columnwidth}{!}{
    \setlength{\tabcolsep}{0.018\linewidth}
    \begin{tabular}{c|c|cccc} \toprule
        Variants & PQ$^\mathrm{Th}$ & Bicycle & Motorcycle & Person & Bicyclist  \\
        \midrule
        Baseline & 63.7 & 52.4 & 59.6 & 77.1 & 90.3 \\
        Ours & \textbf{68.8\textcolor{red}{$_{\uparrow5.1}$}} & \textbf{55.5\textcolor{red}{$_{\uparrow3.1}$}} & \textbf{67.6\textcolor{red}{$_{\uparrow8.0}$}} & \textbf{79.5\textcolor{red}{$_{\uparrow2.4}$}} & \textbf{91.5\textcolor{red}{$_{\uparrow1.2}$}} \\ \bottomrule
    \end{tabular}}
    \label{ab:3}
    \vspace{-1.0em}
\end{table}

\textbf{Effects of Decoupling Strategy.} 
We decouple things/stuff queries while disentangling classification/segmentation to mitigate mutual competition. Table \ref{ab:2} validates the effectiveness of our decoupling strategy.

\emph{1) Effects of decoupling queries:} Baseline employs coupled queries with Hungarian matching and achieves a PQ of 61.3\%. In Variant 1, things/stuff queries are straightforwardly decoupled into independent learnable queries, but this results in only marginal improvements. We argue that the learnable queries, initialized randomly as prototypes, still contain limited information about small instances, making it challenging to converge to specific areas across outdoor scenes.

In Variant 4, decoupled queries generated based on positions and embeddings within the BEV space lead to a substantial gain of 1.8\% and 4.1\% in PQ and PQ$^\mathrm{Th}$, respectively. We attribute this improvement to the alignment of queries with their intrinsic properties, where things queries are extracted from local regions based on BEV positions, while stuff queries encapsulate global context. Additionally, the matching algorithm is replaced by the BEV positions of queries, alleviating competition between things and stuff while promoting individual learning processes. This allows the model to better focus on specific areas of interest, leading to improved segmentation performance.

\emph{2) Effects of decoupling classification/segmentation:} 
In Variants 2 and 3, the query generator assigns semantic classes to queries that are solely responsible for predicting segmentation masks. This results in a 0.3\% and 1.0\% increase in PQ compared to Variant 1. The decoupling strategy diminishes query similarities between different objects of the same class, helping the model to better distinguish between instances.

\textbf{Effects on Segmenting Small Instances.} 
Table \ref{ab:3} presents a detailed comparison of Panoptic Quality for small instances between DQFormer and the baseline model. The results illustrate a significant improvement in these small instances, confirming that our decoupling strategy effectively addresses the mutual competition between objects and large background elements, particularly in the segmentation of smaller instances.

\begin{table}[t]
    \caption{\textbf{Ablation on the similarity thresholds $\theta_{th}$ for things queries fusion} on the SemanticKITTI validation set. Semantic-aware denotes only fusing queries with the same semantics. All scores are in [\%].}
    \vspace{-.5em}
    \centering
    \small
    
    \resizebox{0.95\columnwidth}{!}{
    \setlength{\tabcolsep}{0.02\linewidth}
    \begin{tabular}{cc|ccc|ccc} \toprule
        Semantic-aware & $\theta_{th}$ &  PQ & RQ & SQ & PQ$^\mathrm{Th}$ & RQ$^\mathrm{Th}$ & SQ$^\mathrm{Th}$ \\
     \midrule
        \cmark & 1.00 & 62.8 & 72.6 & 78.5 & 67.3 & 73.1 & 85.4\\
        \cmark & 0.95 & 63.1 & 72.9 & 81.5 & 67.9 & 73.7 & 92.5\\
        \cmark & 0.90 & 63.2 & 73.0 & 81.6 & 68.2 & 74.0 & 92.6\\
        \rowcolor{gray!10} \cmark & 0.85 & \textbf{63.5} & \textbf{73.1} & \textbf{81.7} & \textbf{68.8} & \textbf{74.2} & \textbf{92.9}\\   
        \cmark & 0.80 & 63.4 & \textbf{73.1} & 81.6 & 68.6 & \textbf{74.2} & 92.7\\
        \midrule
        \xmark & 0.85 & 63.3 & 72.9 & 78.8 & 68.3 & 73.6 & 86.1\\
        \bottomrule
    \end{tabular}}
    \label{ab:4}
    % \vspace{-1.5em}
\end{table}

\begin{table}[t]
    \caption{\textbf{Ablation on query fusion and mask fusion modules} on the SemanticKITTI validation set. All scores are in [\%].}
    \vspace{-.5em}
    \centering
    \small
    \resizebox{0.95\columnwidth}{!}{
    \setlength{\tabcolsep}{0.02\linewidth}
    \begin{tabular}{cc|ccc|ccc} \toprule
        query fusion & mask fusion &  PQ & RQ & SQ & PQ$^\mathrm{Th}$ & RQ$^\mathrm{Th}$ & SQ$^\mathrm{Th}$ \\
        \midrule
        \xmark & \xmark & 62.5 & 72.0 & 76.6 & 66.4 & 72.0 & 80.4 \\ 
        \xmark & \cmark & 62.8 & 72.6 & 78.5 & 67.3 & 73.1 & 85.4\\
        \cmark & \xmark & 63.1 & 72.9 & 81.5 & 67.8 & 73.7 & 92.4\\
        \rowcolor{gray!10} \cmark & \cmark & \textbf{63.5} & \textbf{73.1} & \textbf{81.7} & \textbf{68.8} & \textbf{74.2} & \textbf{92.9}\\
        \bottomrule
    \end{tabular}}
    \label{ab:5}
    \vspace{-1.5em}
\end{table}

\textbf{Effects of Query Fusion and Mask Fusion.} 
Table \ref{ab:4} explores the impact of varying the similarity threshold for fusing things queries, revealing optimal performance at $\theta_{th}=0.85$ with semantic-aware fusion. Conversely, employing semantic-agnostic fusion leads to a 6.8\% decrease in SQ$^\mathrm{Th}$ due to the risk of merging instances of different semantics.

Moreover, Table \ref{ab:5} compares our proposed query fusion module with the mask fusion module introduced in \cite{mei2023centerlps}. These modules serve as the pre-fusion approach for integrating queries and the post-fusion method for merging duplicated instance masks, respectively. The results illustrate that combining both fusion techniques yields optimal performance.

\begin{table}[t]
    \caption{\textbf{Ablation on the number of sample points in mask loss} on the nuScenes validation set. $S_{th}$ and $S_{all}$ represent the number of sample points for instances and the entire scene, respectively. All scores are in [\%].}
    \vspace{-.5em}
    \centering
    \small
    
    \resizebox{0.95\columnwidth}{!}{
    \setlength{\tabcolsep}{0.009\linewidth}
    \begin{tabular}{ccc|ccc|ccc} \toprule
        Sub-Sample & $S_{th}$ & $S_{all}$ &  PQ & RQ & SQ & PQ$^\mathrm{Th}$ & RQ$^\mathrm{Th}$ & SQ$^\mathrm{Th}$ \\
        \midrule
        \xmark & - & - & 77.2 & 88.8 & 86.5 & 77.1 & 89.0 & 86.3 \\
        \midrule
        \cmark & 100 & 20,000 & 76.7 & 88.4 & 86.4 & 76.4 & 88.3 & 86.3 \\
        \cmark & 500 & 20,000 & 77.4 & 88.8 & \textbf{86.8} & 77.4 & 88.9 & \textbf{86.8} \\
        \rowcolor{gray!10} \cmark & 1,000 & 20,000 & \textbf{77.7} & \textbf{89.2} & \textbf{86.8} & \textbf{77.8} & \textbf{89.5} & 86.7 \\
        \cmark & 500 & 10,000 & 77.0 & 88.7 & 86.4 & 76.8 & 88.8 & 86.2 \\
        \cmark & 1,000 & 10,000 & 77.3 & 88.9 & 86.6 & 77.4 & 89.1 & 86.5 \\
        \bottomrule
    \end{tabular}}
    \label{ab:6}
    \vspace{-1.0em}
\end{table}

\begin{table}[t]
    \caption{\textbf{Ablation on the mask loss functions} on the nuScenes validation set. BCE denotes the binary cross-entropy loss. All scores are in [\%].}
    \vspace{-.5em}
    \centering
    \small
    
    \resizebox{0.95\columnwidth}{!}{
    \setlength{\tabcolsep}{0.02\linewidth}
    \begin{tabular}{ccc|ccc|ccc} \toprule
        BCE & Focal & Dice &  PQ & RQ & SQ & PQ$^\mathrm{Th}$ & RQ$^\mathrm{Th}$ & SQ$^\mathrm{Th}$ \\
        \midrule
        \rowcolor{gray!10} \cmark & - & \cmark & \textbf{77.7} & \textbf{89.2} & \textbf{86.8} & \textbf{77.8} & \textbf{89.5} & \textbf{86.7} \\
        \cmark & \cmark & - & 76.4 & 87.8 & 86.5 & 76.0 & 87.6 & 86.4 \\
        - & \cmark & \cmark & 77.1 & 88.7 & 86.5 & 77.0 & 89.0 & 86.3 \\
        \cmark & \cmark & \cmark & 77.4 & 89.0 & 86.6 & 77.5 & 89.3 & 86.5 \\
        \bottomrule
    \end{tabular}}
    \label{ab:7}
    \vspace{-1.5em}
\end{table}

\textbf{Effects of Mask Loss.}
As previously noted, instances are typically concentrated within specific local regions and are significantly smaller than the entire scene. Consequently, the imbalance between positive and negative samples in mask supervision becomes critical for instances.

Drawing from insights in prior studies \cite{cheng2021per, marcuzzi2023mask}, we employ a strategy that randomly samples a fixed number of $S_{th}$ points from objects and $S_{all}$ points from the entire scene to calculate the mask loss, thus achieving a more balanced mask supervision. 
Ablation studies in Table \ref{ab:6} demonstrate the effectiveness of this sub-sampling strategy and the impact of the number of sample points on mask loss. Without the sub-sampling strategy, segmentation quality for objects (SQ$^\mathrm{Th}$) suffers the most. Moreover, using a limited number of sample points for objects results in inferior performance in PQ$^\mathrm{Th}$. The optimal configuration is found with $S_{th}=1000$ and $S_{all}=20000$, achieving a more balanced distribution for objects and providing more points for mask supervision.

Additionally, Table \ref{ab:7} compares the performances of different loss functions for mask supervision, including Binary Cross-Entropy loss (BCE), Focal loss, and Dice loss. The results indicate the significant contribution of Dice loss in improving the panoptic quality of objects (PQ$^\mathrm{Th}$). The combination of BCE and Dice loss produces the most favorable outcomes.

\begin{figure*}[t]
\centering
	\includegraphics[width=0.98\textwidth]{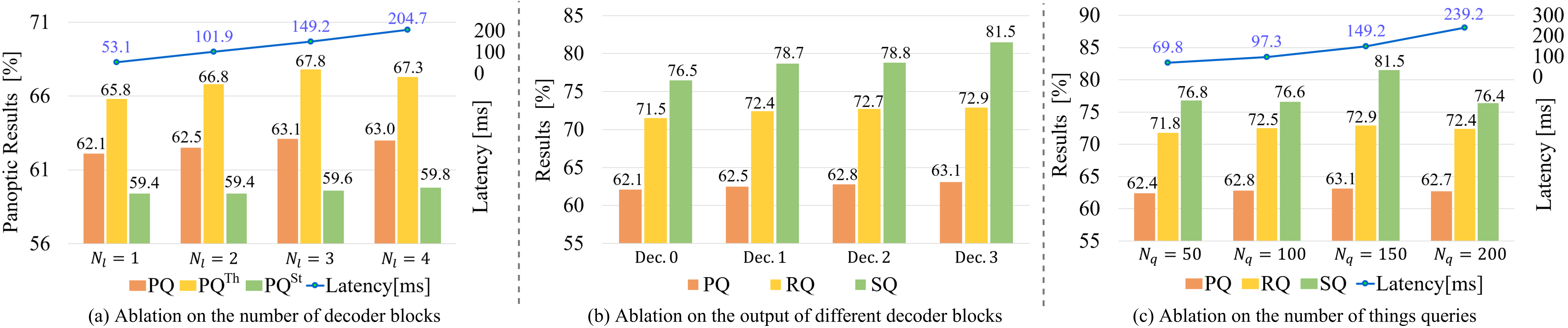}
        \vspace{-.5em}
	\caption{\textbf{Ablation on the number of decoder blocks and things queries} on the SemanticKITTI validation set, along with the latency of mask decoding.}
	\label{ab:8}
        \vspace{-.5em}
\end{figure*}

\begin{table*}[t]
    \caption{\textbf{Ablation on model settings and efficiency} between DQFormer and existing methods. Here, $N_L$ represents the number of decoder blocks, $N_q$ denotes the instance query number, and $C_e$ indicates the feature dimension. All experiments are conducted on the nuScenes validation split. \\ $\dag$ indicates that we measure the latency on our hardware using the officially released codes. }
    \vspace{-.5em}
    \centering
    \small
    \setlength{\tabcolsep}{0.014\linewidth}
    \scalebox{0.88}{
    \begin{tabular}{l|ccc|ccccc|cc} 
        \toprule
        Method & $N_L$ & $N_q$ & $C_e$ & PQ & RQ& SQ & PQ$^\mathrm{Th}$ & PQ$^\mathrm{St}$ & Params & FPS \\
        \midrule
        % \midrule
        EfficientLPS \cite{sirohi2021efficientlps} & - & - & - & 62.0 & 73.9 & 83.4 & 56.8 & 70.6 & 43.8M & 4.0$^\dag$\\ %4.01
        DS-Net \cite{hong2021lidar} & - & - & - & 42.5 & 50.3 & 83.6 & 32.5 & 59.2 & 56.5M & 2.2$^\dag$ \\ %2.24
        MaskPLS-M \cite{marcuzzi2023mask} & - & - & - & 57.7 & 66.0 & 71.8 & 64.4 & 52.2 & 31.5M & 4.7$^\dag$\\  %4.72
        \midrule
        
        \multirow{4}{*}{\makecell{DQFormer (Ours)}} & 1 & 150 & 128 & 76.2 & 88.1 & 86.1 & 75.7 & 77.1 & 41.9M & \textbf{5.3}  \\
         & 2 & 150 & 256 & 76.9 & 88.9 & 86.2 & 76.6 & \textbf{77.6} & 47.3M & 4.6 \\
         & \cellcolor{gray!10}3 & \cellcolor{gray!10}150 & \cellcolor{gray!10}256 & \cellcolor{gray!10}\textbf{77.7} & \cellcolor{gray!10}\textbf{89.2} & \cellcolor{gray!10}\textbf{86.8} & \cellcolor{gray!10}\textbf{77.8} & \cellcolor{gray!10}77.5 & \cellcolor{gray!10}50.5M & \cellcolor{gray!10}4.5 \\ 
         & 3 & 200 & 256 & 77.0 & 88.8 & 86.3 & 76.6 & 77.5 & 50.5M & 4.4 \\
         \bottomrule
    \end{tabular}}
    \vspace{-.5em}
    \label{ab:9}
\end{table*}

\textbf{Effects of the Decoder Blocks and Things Queries.} 
In Figure \ref{ab:8}, we compare various settings involving the number of mask decoder blocks and object queries, along with the latency of mask decoding calculated on a single RTX A6000 GPU.

In Figure \ref{ab:8}(a), it is evident that employing a greater number of decoder blocks leads to enhancements in both PQ and PQ$^\mathrm{Th}$. This improvement can be attributed to the masked cross-attention mechanism combined with deep supervision, which directs the queries to attend to specific key points within each block, progressively refining the segmentation quality.

In Figure \ref{ab:8}(b), we present the evaluation of each block's performance. The results indicate that deeper blocks exhibit superior performance, suggesting a gradual convergence toward keypoints facilitated by the masked attention mechanism.

In Figure \ref{ab:8}(c), we examine the effect of the number of object queries $N_q$. Optimal performance in terms of both PQ and SQ is observed when $N_q=150$. We explain that using too few queries may overlook potential objects that are challenging to discover 
Conversely, an excessive number of queries leads to over-segmentation issues and escalates computational costs.

\textbf{Effects of Model Settings and Efficiency.}
Table \ref{ab:9} presents an evaluation of the performance and efficiency of our DQFormer in comparison to existing methods. 
We compare our model with three representative approaches: detection-based EfficientLPS \cite{sirohi2021efficientlps}, clustering-based DSNet \cite{hong2021lidar}, and query-based MaskPLS \cite{marcuzzi2023mask}. The running speeds are computed on the same NVIDIA RTX A6000 GPU on the nuScenes validation set.

The results reveal a significant performance enhancement with our DQFormer, surpassing previous methods by a substantial margin. Additionally, our approach exhibits a faster running speed compared to detection-based and clustering-based methods while comparable to the query-based MaskPLS. This underscores the efficiency of our unified workflow, which eliminates time-consuming detection processes or clustering algorithms, thereby enhancing overall efficiency.

Comparisons across different settings of DQFormer demonstrate that heavier network configurations, \ie deeper decoder blocks and larger feature dimensions, lead to improved performance. However, they also come with increased computational costs and reduced inference speed. To strike a balance between accuracy and efficiency, we establish $N_l=3$ decoder blocks and $C_e=256$ feature dimension as the default configuration.

\subsection{Qualitative Results and Discussion}
In this section, we present visualization results, including qualitative comparisons, mask predictions across decoder blocks, object center predictions, and attention maps of things/stuff queries.

\begin{figure*}[t]
\centering
	\includegraphics[width=0.92\textwidth]{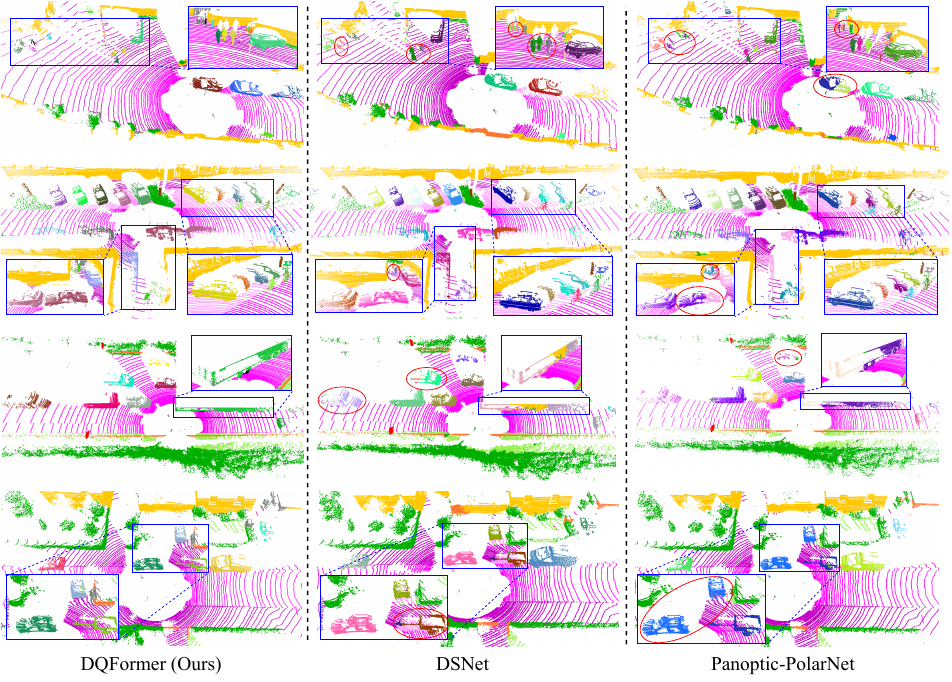}
        \vspace{-1em}
	\caption{\centering{\textbf{Qualitative comparisons of panoptic segmentation} between our DQFormer with DSNet \cite{hong2021lidar} and Panoptic-PolarNet \cite{zhou2021panoptic}, on the SemanticKITTI test split.}}
	\label{fig:vis_comp}
\end{figure*}

\textbf{Comparisons of Panoptic Segmentation.}
Figure \ref{fig:vis_comp} illustrates qualitative comparisons between our DQFormer and two existing methods, DSNet \cite{hong2021lidar} and Panoptic-PolarNet \cite{zhou2021panoptic}, using the SemanticKITTI test set.

The first two rows demonstrate the superior ability of our DQFormer to accurately segment small instances concentrated in local regions. In contrast, DSNet and Panoptic-PolarNet encounter under-segmentation challenges, particularly when dealing with adjacent instances that share similar geometric features. This highlights the effectiveness of our query generator in distinguishing individual objects and generating informative queries for decoding segmentation masks.

In the third row, our DQFormer efficiently segments large objects such as buses or trucks. However, the other two clustering-based methods face over-segmentation issues, especially when handling sparse instances with scattered points.
The last row showcases DQFormer's accuracy in identifying and segmenting rare objects, such as the trolley. These results also underscore our method's proficiency in accurately distinguishing widely distributed stuff points based on their distinct attributes.

\begin{figure*}[t]
\centering
	\includegraphics[width=0.88\textwidth]{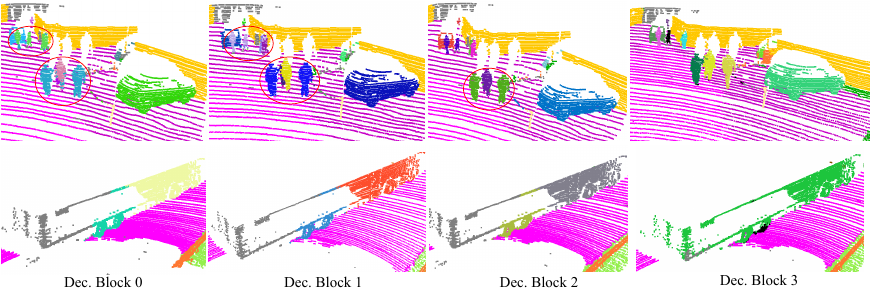}
        \vspace{-1em}
	\caption{\centering{\textbf{Qualitative comparisons of mask predictions} across different decoder blocks on the SemanticKITTI test split.}}
	\label{fig:vis_dec}
    \vspace{-1em}
\end{figure*}

\textbf{Comparisons Across Decoder Blocks.}
In Figure \ref{fig:vis_dec}, we provide visualizations of mask predictions across various decoder blocks. It is evident that shallow blocks encounter under-segmentation issues when dealing with adjacent small targets, and they also yield fragmented masks for large objects. In contrast, deeper blocks demonstrate the ability to generate more precise masks for both objects (things) and background elements (stuff). This observation highlights the effectiveness of the masked cross-attention mechanism, where queries progressively focus on key points guided by previous masks, ultimately leading to more accurate segmentation predictions.

\begin{figure}[t]
\centering
	\includegraphics[width=0.48\textwidth]{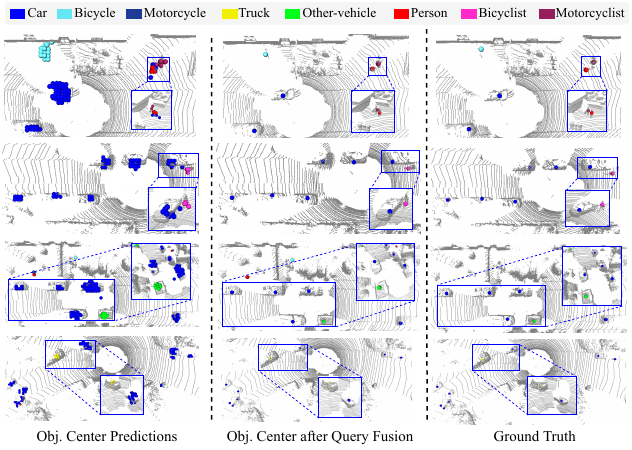}
        \vspace{-1em}
	\caption{\textbf{Visualization of object centers} extracted from the BEV heatmap, after the query fusion module, alongside the corresponding ground truth.}
	\label{fig:vis_center}
    % \vspace{-.5em}
\end{figure}

\textbf{Visualization of the Object Centers.}
In Figure \ref{fig:vis_center}, we present visualizations of predicted object centers, after the query fusion module, alongside their corresponding ground truth. We use distinct colors to represent different predicted categories.

Upon comparing the first and second columns, the results demonstrate the efficiency of our query fusion module in merging duplicated queries associated with the same objects. This merging process results in compact queries that closely align with the ground truth spatially.

The first two rows illustrate that our query generator adept at localizing adjacent small instances and seamlessly fusing their individual queries. 
This accurate fusion is achieved through two constraints: merging queries within the same small window for geometric consistency and utilizing query similarity to maintain feature constraints in the embedding space.

The results of the third and fourth rows highlight the query generator's ability to distinguish adjacent instances sharing similar geometric attributes accurately. This capability enables the generation of object-specific queries, facilitating decoding corresponding masks. 

In the last row, we observe a failure case in our query fusion module, where all queries related to a \emph{truck} are not fused into a single query embedding. This emphasizes the importance of the mask fusion process, showcasing its indispensability in effectively merging duplicated masks.

\begin{figure}[t]
\centering
	\includegraphics[width=0.48\textwidth]{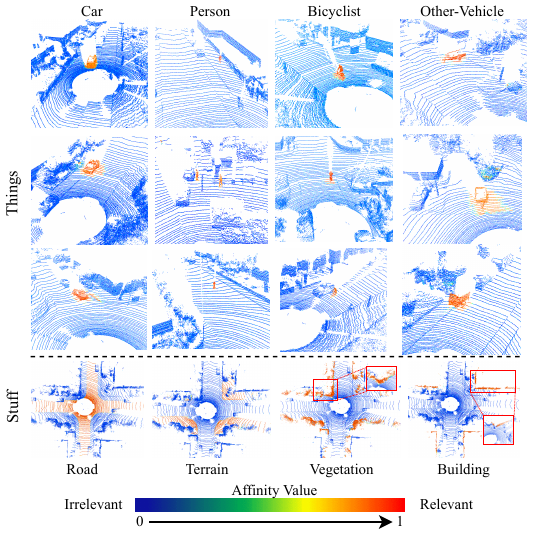}
        \vspace{-1em}
	\caption{\textbf{Visualization of attention maps} showcasing the affinities between things/stuff queries and the point cloud.}
	\label{fig:vis_attention}
    \vspace{-.5em}
\end{figure}

\textbf{Visualization of Attention Maps.}
In Figure \ref{fig:vis_attention}, we explore the relationship between things/stuff queries and the point cloud, with red indicating higher correlations and blue indicating lower ones. The visualization demonstrates that our queries exhibit high alignment with specific areas of interest across various categories. Specifically, things queries focus on locally concentrated points, while stuff queries attend to points distributed throughout the scene. This underscores the ability of query embeddings to capture practical features, aiding in the generation of precise segmentation masks via the masked-attention mechanism.

\section{Conclusion}
We propose a novel framework named DQFormer for unified LiDAR panoptic segmentation with decoupled queries, aiming to address the challenges of mutual competition between things/stuff and the ambiguity between classification/segmentation. Specifically, we introduce a multi-scale query generator that proposes semantic-aware queries based on the positions and embeddings of things/stuff in BEV space. Moreover, we design a query fusion module to integrate queries from multiple BEV resolutions. Finally, we propose a query-oriented mask decoder to guide the segmentation process using informative queries. Comprehensive experiments on nuScenes and SemanticKITTI datasets demonstrate that our DQFormer achieves state-of-the-art performance for LiDAR panoptic segmentation. Extensive ablation studies and visualization results further demonstrate the effectiveness of our method.

\bibliographystyle{IEEEtran}
\bibliography{main}

\end{document}